\documentclass[10pt,twocolumn,letterpaper]{article}

\usepackage{wacv}
\usepackage{times}
\usepackage{epsfig}
\usepackage{graphicx}
\usepackage{amsmath}
\usepackage{amssymb}

\usepackage[dvipsnames]{xcolor}
\usepackage{tabularx}
\usepackage{multirow,makecell}
\usepackage{booktabs}
\usepackage{float}
\usepackage[font=small]{caption}
\usepackage{subcaption}
\usepackage{grid-system}
\usepackage[ruled,vlined]{algorithm2e}

\def\wacvPaperID{96} % Enter the WACV Paper ID here

\wacvfinalcopy % *** Uncomment this line for the final submission

\ifwacvfinal
\def\assignedStartPage{1} % *** Enter the assigned starting page number (instead of 1)
\fi

\ifwacvfinal
\usepackage[breaklinks=true,bookmarks=false]{hyperref}
\else
\usepackage[pagebackref=true,breaklinks=true,colorlinks,bookmarks=false]{hyperref}
\fi

\ifwacvfinal
\else
\fi

\begin{document}

%%%%%%%%% TITLE
\title{Robust High-Resolution Video Matting with Temporal Guidance}

\author{
Shanchuan Lin$^1$\thanks{Work performed during an internship at ByteDance.} \quad Linjie Yang$^2$ \quad Imran Saleemi$^2$ \quad Soumyadip Sengupta$^1$ \\
$^1$University of Washington \quad $^2$ByteDance Inc. \\

{\tt\small \{linsh,soumya91\}@cs.washington.edu} \quad {\tt\small \{linjie.yang,imran.saleemi\}@bytedance.com}
}

\maketitle
%\thispagestyle{empty}

%%%%%%%%% UTILITIES

\newif\ifcomments
\commentstrue % COMMENT THIS LINE TO DISABLE COMMENTS
\ifcomments
    \newcommand{\peter}[1]{\textcolor{orange}{[Peter: #1]}}
    \newcommand{\soumyadip}[1]{\textcolor{red}{[SS: #1]}}
    \newcommand{\imran}[1]{\textcolor{blue}{[IS: #1]}}
\else
    \providecommand{\peter}[1]{}
    \providecommand{\soumyadip}[1]{}
    \providecommand{\imran}[1]{}
\fi

\begin{abstract}
    \vspace{-5pt}
    We introduce a robust, real-time, high-resolution human video matting method that achieves new state-of-the-art performance. Our method is much lighter than previous approaches and can process 4K at 76 FPS and HD at 104 FPS on an Nvidia GTX 1080Ti GPU. Unlike most existing methods that perform video matting frame-by-frame as independent images, our method uses a recurrent architecture to exploit temporal information in videos and achieves significant improvements in temporal coherence and matting quality. Furthermore, we propose a novel training strategy that enforces our network on both matting and segmentation objectives. This significantly improves our model's robustness. Our method does not require any auxiliary inputs such as a trimap or a pre-captured background image, so it can be widely applied to existing human matting applications. Our code is available at \small{\url{https://peterl1n.github.io/RobustVideoMatting/}}
\end{abstract}
\vspace{-15pt}
\section{Introduction}

Matting is the process of predicting the alpha matte and foreground color from an input frame. Formally, a frame $I$ can be viewed as the linear combination of a foreground $F$ and a background $B$ through an $\alpha$ coefficient:
\vspace{-5pt}
\begin{equation}
    I = \alpha F + (1 - \alpha) B
    \vspace{-5pt}
\end{equation}

\noindent
By extracting $\alpha$ and $F$, we can composite the foreground object to a new background, achieving the background replacement effect.

Background replacement has many practical applications. Many rising use cases, \eg video conferencing and entertainment video creation, need real-time background replacement on human subjects without green-screen props. Neural models are used for this challenging problem but the current solutions are not always robust and often generate artifacts. Our research focuses on improving the matting quality and robustness for such applications.

Most existing methods \cite{modnet,bgmv2,bgmv1}, despite being designed for video applications, process individual frames as independent images. Those approaches neglect the most widely available feature in videos: temporal information. Temporal information can improve video matting performance for many reasons. First, it allows the prediction of more coherent results, as the model can see multiple frames and its own predictions. This significantly reduces flicker and improves perceptual quality. Second, temporal information can improve matting robustness. In the cases where an individual frame might be ambiguous, \eg the foreground color becomes similar to a passing object in the background, the model can better guess the boundary by referring to the previous frames. Third, temporal information allows the model to learn more about the background over time. When the camera moves, the background behind the subjects is revealed due to the perspective change. Even if the camera is fixed, the occluded background still often reveals due to the subject's movements. Having a better understanding of the background simplifies the matting task. Therefore, we propose a recurrent architecture to exploit the temporal information. Our method significantly improves the matting quality and temporal coherence. It can be applied to all videos without any requirements for auxiliary inputs, such as a manually annotated trimap or a pre-captured background image.

Furthermore, we propose a new training strategy to enforce our model on both matting and semantic segmentation objectives simultaneously. Most existing methods \cite{modnet,bgmv2,bgmv1} are trained on synthetic matting datasets. The samples often look fake and prevent the network to generalize to real images. Previous works \cite{modnet,bgmv2} have attempted to initialize the model with weights trained on segmentation tasks, but the model still overfits to the synthetic distribution during the matting training. Others have attempted adversarial training \cite{bgmv1} or semi-supervised learning \cite{modnet} on unlabeled real images as an additional adaptation step. We argue that human matting tasks are closely related to human segmentation tasks. Simultaneously training with a segmentation objective can effectively regulate our model without additional adaptation steps.

Our method outperforms the previous state-of-the-art method while being much lighter and faster. Our model uses only 58\% parameters and can process real-time high-resolution videos at 4K 76 FPS and HD 104 FPS on an Nvidia GTX 1080Ti GPU.
\section{Related Works}

\textbf{Trimap-based matting.} Classical (non-learning) algorithms \cite{aksoy2017designing,chen2013knn,chuang2001bayesian,gastal2010shared,levin2007closed,levin2008spectral,sun2004poisson} require a manual trimap annotation to solve for the unknown regions of the trimap. Such methods are reviewed in the survey by Wang and Cohen \cite{mattingsurvey}. Xu \etal \cite{deepimagematting} first used a deep network for trimap-based matting and many recent research continued this approach. FBA \cite{fbamatting} is one of the latest. Trimap-based methods are often object agnostic (not limited to human). They are suitable for interactive photo-editing applications where the user can select target objects and provide manual guidance. To extend it to video, Sun \etal proposed DVM \cite{dvm}, which only requires a trimap on the first frame and can propagate it to the rest of the video.

\textbf{Background-based matting.} Soumyadip \etal proposed background matting (BGM) \cite{bgmv1}, which requires an additional pre-captured background image as input. This information acts as an implicit way for foreground selection and increases matting accuracy. Lin and Ryabtsev \etal further proposed BGMv2 \cite{bgmv2} with improved performance and a focus on real-time high-resolution. However, background matting cannot handle dynamic backgrounds and large camera movements.

\textbf{Segmentation.} Semantic segmentation is to predict a class label for every pixel, often without auxiliary inputs. Its binary segmentation mask can be used to locate the human subjects, but using it directly for background replacement will result in strong artifacts. Nonetheless, segmentation tasks are similar to matting tasks in an auxiliary-free setting, and the research in segmentation inspires our network design. DeepLabV3 \cite{deeplabv3} proposed ASPP (Atrous Spatial Pyramid Pooling) module and used dilated convolution in its encoder to improve performance. This design has been adopted by many following works, including MobileNetV3 \cite{mobilenetv3}, which simplified ASPP to LR-ASPP.

\textbf{Auxiliary-free matting.} Fully automatic matting without any auxiliary inputs has also been studied. Methods like \cite{qiao2020attention,zhang2019late} work on any foreground objects but not as robust, while others like \cite{modnet,shen2016deep,zhu2017fast} are trained specifically for human portraits. MODNet \cite{modnet} is the latest portrait matting method. In contrast, our method is trained to work well on the full human body.

\textbf{Video matting.} Very few neural matting methods are designed for videos natively. MODNet \cite{modnet} proposed a post-processing trick that compares the prediction of neighboring frames to suppress flicker, but it cannot handle fast-moving body-parts and the model itself still operates on frames as independent images. BGM \cite{bgmv1} explored taking a few neighboring frames as additional input channels, but this only provides short-term temporal cues and its effects were not the focus of the study. DVM \cite{deepimagematting} is video-native but focused on utilizing temporal information to propagate trimap annotations. On contrary, our method focuses on using temporal information to improve matting quality in an auxiliary-free setting.

\textbf{Recurrent architecture.} Recurrent neural network has been widely used for sequence tasks. Two of the most popular architectures are LSTM (Long Short-Term Memory) \cite{lstm} and GRU (Gated Recurrent Unit) \cite{gru}, which have also being adopted to vision tasks as ConvLSTM \cite{convlstm} and ConvGRU \cite{convgru}. Previous works have explored using recurrent architectures for various video vision tasks and showed improved performance compared to the image-based counterparts \cite{Ventura2019RVOSER,Pfeuffer2019SemanticSO,Tokmakov2017LearningVO}. Our work adopts recurrent architectures to the matting task.

\textbf{High-resolution matting.} Patch-based refinement has been explored by PointRend \cite{pointrend} for segmentation and BGMv2 \cite{bgmv2} for matting. It only performs convolution on selective patches. Another approach is using Guided Filter \cite{guidedfilter}, a post-processing filter that jointly upsamples the low-resolution prediction given the high-resolution frame as guidance. Deep Guided Filter (DGF) \cite{deepguidedfilter} was proposed as a learnable module that can be trained with the network end-to-end without manual hyperparameters. Despite filter-based upsampling being less powerful, we choose it because it is faster and well supported by all inference frameworks.
\section{Model Architecture}

\begin{figure*}
    \centering
    \includegraphics[width=\textwidth]{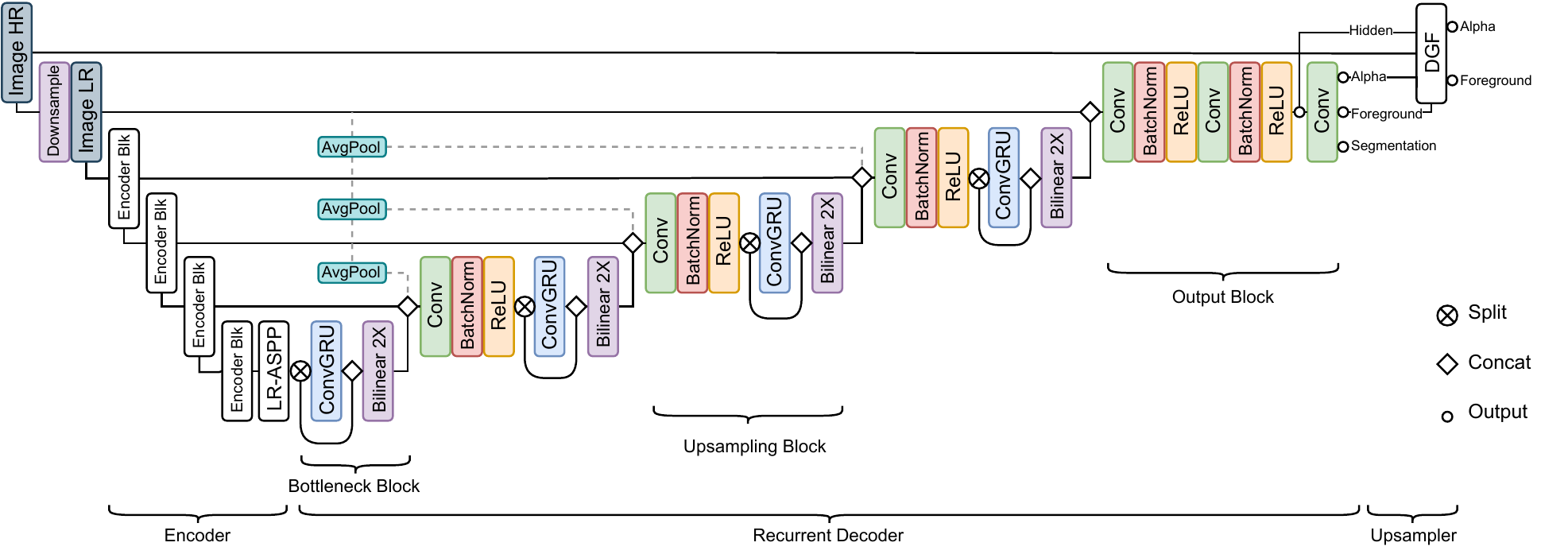}
    \vspace{-20pt}
    \caption{Our network consists of a feature-extraction encoder, a recurrent decoder, and Deep Guided Filter (DGF) module. To process high-resolution videos, the input is first downsampled for the encoder-decoder network, then DGF is used to upsample the result.}
    \label{fig:model_architecture}
    \vspace{-10pt}
\end{figure*}

Our architecture consists of an encoder that extracts individual frame's features, a recurrent decoder that aggregates temporal information, and a Deep Guided Filter module for high-resolution upsampling. Figure \ref{fig:model_architecture} shows our model architecture.

\subsection{Feature-Extraction Encoder}

Our encoder module follows the design of state-of-the-art semantic segmentation networks \cite{deeplabv3,deeplabv3+,mobilenetv3} because the ability to accurately locate human subjects is fundamental to the matting task. We adopt MobileNetV3-Large \cite{mobilenetv3} as our efficient backbone followed by the LR-ASPP module as proposed by MobileNetV3 for semantic segmentation tasks. Noticeably, the last block of MobileNetV3 uses dilated convolutions without downsampling stride. The encoder module operates on individual frames and extracts features at $\frac{1}{2}$, $\frac{1}{4}$, $\frac{1}{8}$, and $\frac{1}{16}$ scales for the recurrent decoder.

\subsection{Recurrent Decoder}

We decide to use a recurrent architecture instead of attention or simply feedforwarding multiple frames as additional input channels for several reasons. Recurrent mechanisms can learn what information to keep and forget by itself on a continuous stream of video, while the other two methods must rely on a fixed rule to remove old and insert new information to the limited memory pool on every set interval. The ability to adaptively keep both long-term and short-term temporal information makes recurrent mechanisms more suitable for our task.

Our decoder adopts ConvGRU at multiple scales to aggregate temporal information. We choose ConvGRU because it is more parameter efficient than ConvLSTM by having fewer gates. Formally, ConvGRU is defined as:

\vspace{-15pt}
\begin{equation}
    \begin{aligned}
        & z_t = \sigma (w_{zx} * x_t + w_{zh} * h_{t-1} + b_z) \\
        & r_t = \sigma (w_{rx} * x_t + w_{rh} * h_{t-1} + b_r) \\
        & o_t = tanh(w_{ox} * x_t + w_{oh} * (r_t \odot h_{t-1}) + b_o) \\
        & h_t = z_t \odot h_{t-1} + (1 - z_t) \odot o_t
    \end{aligned}
    \vspace{-5pt}
\end{equation}

\noindent
where operators $*$ and $\odot$ represent convolution and element-wise product respectively; $tanh$ and $\sigma$ represent hyperbolic tangent and sigmoid function respectively. $w$ and $b$ are the convolution kernel and the bias term. The hidden state $h_t$ is used as both the output and the recurrent state to the next time step as $h_{t-1}$. The initial recurrent state $h_0$ is an all zero tensor.

As shown in Figure \ref{fig:model_architecture}, our decoder consists of a bottleneck block, upsampling blocks, and an output block.

\textbf{Bottleneck block} operates at the $\frac{1}{16}$ feature scale after the LR-ASPP module. A ConvGRU layer is operated on only half of the channels by split and concatenation. This significantly reduces parameters and computation since ConvGRU is computationally expansive.

\textbf{Upsampling block} is repeated at $\frac{1}{8}$, $\frac{1}{4}$, and $\frac{1}{2}$ scale. First, it concatenates the bilinearly upsampled output from the previous block, the feature map of the corresponding scale from the encoder, and the input image downsampled by repeated $2\times2$ average pooling. Then, a convolution followed by Batch Normalization \cite{batchnorm} and ReLU \cite{relu} activation is applied to perform feature merging and channel reduction. Finally, a ConvGRU is applied to half of the channels by split and concatenation.

\textbf{Output block} does not use ConvGRU because we find it expansive and not impactful at this scale. The block only uses regular convolutions to refine the results. It first concatenates the input image and the bilinearly upsampled output from the previous block. Then it employs 2 repeated convolution, Batch Normalization, and ReLU stacks to produce the final hidden features. Finally, the features are projected to outputs, including 1-channel alpha prediction, 3-channel foreground prediction, and 1-channel segmentation prediction. The segmentation output is used for the segmentation training objective as later described in Section \ref{sec:training}.

We find applying ConvGRU on half of the channels by split and concatenation effective and efficient. This design helps ConvGRU to focus on aggregating temporal information, while the other split branch forwards the spatial features specific to the current frame. All convolutions use $3\times3$ kernels, except the last projection uses a $1\times1$ kernel.

We modify our network such that it can be given $T$ frames at once as input and each layer processes all $T$ frames before passing to the next layer. During training, this allows Batch Normalization to compute statistics across both batch and time to ensure the normalization is consistent. During inference, $T=1$ can be used to process live videos and $T>1$ can be used to exploit more GPU parallelism from the non-recurrent layers as a form of batching if the frames are allowed to be buffered. Our recurrent decoder is uni-directional so it can be used for both live streaming and post-processing.

\subsection{Deep Guided Filter Module}
\label{sec:model_refinement}

We adopt Deep Guided Filter (DGF) as proposed in \cite{deepguidedfilter} for high-resolution prediction. When processing high-resolution videos such as 4K and HD, we downsample the input frame by a factor $s$ before passing through the encoder-decoder network. Then the low-resolution alpha, foreground, final hidden features, as well as the high-resolution input frame are provided to the DGF module to produce high-resolution alpha and foreground. The entire network is trained end-to-end as described in Section \ref{sec:training}. Note that the DGF module is optional and the encoder-decoder network can operate standalone if the video to be processed is low in resolution.

% More details are in the supplementary.

Our entire network does not use any special operators and can be deployed to most existing inference frameworks. More architectural details are in the supplementary.
\section{Training}
\label{sec:training}

We propose to train our network with both matting and semantic segmentation objectives simultaneously for several reasons:

First, human matting tasks are closely related to human segmentation tasks. Unlike trimap-based and background-based matting methods that are given additional cues as inputs, our network must learn to semantically understand the scene and be robust in locating the human subjects.

Second, most existing matting datasets only provide ground-truth alpha and foreground that must be synthetically composited to background images. The compositions sometimes look fake due to the foreground and the background having different lighting. On the other hand, semantic segmentation datasets feature real images where the human subjects are included in all types of complex scenes. Training with semantic segmentation datasets prevents our model from overfitting to the synthetic distribution.

Third, there is a lot more training data available for semantic segmentation tasks. We harvest a variety of publically available datasets, both video-based and image-based, to train a robust model.

\subsection{Matting Datasets}

Our model is trained on VideoMatte240K (VM) \cite{bgmv2}, Distinctions-646 (D646) \cite{hamatting}, and Adobe Image Matting (AIM) \cite{deepimagematting} datasets. VM provides 484 4K/HD video clips. We divide the dataset into 475/4/5 clips for train/val/test splits. D646 and AIM are image matting datasets. We use only images of humans and combine them to form 420/15 train/val splits for training. For evaluation, D646 and AIM each provide 11 and 10 test images respectively.

For backgrounds, the dataset by \cite{dvm} provides HD background videos that are suitable for matting composition. The videos include a variety of motions, such as cars passing, leaves shaking, and camera movements. We select 3118 clips that does not contain humans and extract the first 100 frames from every clip. We also crawl 8000 image backgrounds following the approach of \cite{bgmv2}. The images have more indoor scenes such as offices and living rooms.

We apply motion and temporal augmentations on both foreground and background to increase data variety. Motion augmentations include affine translation, scale, rotation, sheer, brightness, saturation, contrast, hue, noise and blur that change continuously over time. The motion is applied with different easing functions such that the changes are not always linear. The augmentation also adds artificial motions to the image datasets. Additionally, we apply temporal augmentation on videos, including clip reversal, speed changes, random pausing, and frame skipping. Other discrete augmentations \ie horizontal flip, grayscale, and sharpening, are applied consistently to all frames.

\subsection{Segmentation Datasets}

We use video segmentation dataset YouTubeVIS and select 2985 clips containing humans. We also use image segmentation datasets COCO \cite{coco} and SPD \cite{spd}. COCO provides 64,111 images containing humans while SPD provides additional 5711 samples. We apply similar augmentations but without motion, since YouTubeVIS already contains large camera movements and the image segmentation datasets do not require motion augmentation.

\subsection{Procedures}

Our matting training is pipelined into four stages. They are designed to let our network progressively see longer sequences and higher resolutions to save training time. We use Adam optimizer for training. All stages use batch size $B=4$ split across 4 Nvidia V100 32G GPUs.

\textbf{Stage 1:} We first train on VM at low-resolution without the DGF module for 15 epochs. We set a short sequence length $T=15$ frames so that the network can get updated quicker. The MobileNetV3 backbone is initialized with pretrained ImageNet \cite{imagenet} weights and uses $1e^{-4}$ learning rate, while the rest of the network uses $2e^{-4}$. We sample the height and width of the input resolution $h$, $w$ independently between 256 and 512 pixels. This makes our network robust to different resolutions and aspect ratios.

\textbf{Stage 2:} We increase $T$ to 50 frames, reduce the learning rate by half, and keep other settings from stage 1 to train our model for 2 more epochs. This allows our network to see longer sequences and learn long-term dependencies. $T=50$ is the longest we can fit on our GPUs.

\textbf{Stage 3:} We attach the DGF module and train on VM with high-resolution samples for 1 epoch. Since high resolution consumes more GPU memory, the sequence length must be set to very short. To avoid our recurrent network overfitting to very short sequences, we train our network on both low-resolution long sequences and high-resolution short sequences. Specifically, the low-resolution pass does not employ DGF and has $T=40$ and $h,w \sim (256, 512)$. The high-resolution pass entails the low-resolution pass and employs DGF with downsample factor $s=0.25$, $\hat{T}=6$ and $\hat{h},\hat{w} \sim (1024, 2048)$. We set the learning rate of DGF to $2e^{-4}$ and the rest of the network to $1e^{-5}$.

\textbf{Stage 4:} We train on the combined dataset of D646 and AIM for 5 epochs. We increase the decoder learning rate to $5e^{-5}$ to let our network adapt and keep other settings from stage 3.

\textbf{Segmentation:} Our segmentation training is interleaved between every matting training iteration. We train the network on image segmentation data after every odd iteration, and on video segmentation data after every even ones. Segmentation training is applied to all stages. For video segmentation data, we use the same $B$, $T$, $h$, $w$ settings following every matting stage. For image segmentation data, we treat them as video sequences of only 1 frame, so $T'=1$. This gives us room to apply a larger batch size $B'=B \times T$. Since the images are feedforwarded as the first frame, it forces the segmentation to be robust even in the absence of recurrent information.

\subsection{Losses}

We apply losses on all $t \in [1, T]$ frames. To learn alpha $\alpha_t$ \wrt ground-truth $\alpha^*_t$, we use L1 loss $\mathcal{L}^{\alpha}_{l1}$ and pyramid Laplacian loss $\mathcal{L}^{\alpha}_{lap}$, as reported by \cite{fbamatting,contextawarematting} to produce the best result. We also apply a temporal coherence loss $\mathcal{L}^{\alpha}_{tc}$, as used by \cite{dvm}, to reduce flicker:
\vspace{-5pt}
\begin{equation}
    \mathcal{L}^{\alpha}_{l1} = || \alpha_t - \alpha^*_t ||_1
\end{equation}
\begin{equation}
    \mathcal{L}^{\alpha}_{lap} = \sum_{s=1}^{5} \frac{2^{s-1}}{5} ||L^{s}_{pyr}(\alpha_t) - L^{s}_{pyr}(\alpha^*_t) ||_1
\end{equation}
\begin{equation}
    \mathcal{L}^{\alpha}_{tc} = || \frac{\text{d}\alpha_t}{\text{d}t} - \frac{\text{d}\alpha^*_t}{\text{d}t} ||_2
\end{equation}

To learn foreground $F_t$ \wrt ground-truth $F^*_t$, We compute L1 loss $\mathcal{L}^{F}_{l1}$ and temporal coherence loss $\mathcal{L}^{F}_{tc}$ on pixels where $\alpha^*_t > 0$ following the approach of \cite{bgmv2}:
\vspace{-5pt}
\begin{equation}
    \mathcal{L}^{F}_{l1} = || (a^*_t > 0) * (F_t - F^*_t) ||_1
    \vspace{-5pt}
\end{equation}

\begin{equation}
    \mathcal{L}^{F}_{tc} = || (a^*_t > 0) * (\frac{\text{d}F_t}{\text{d}t} - \frac{\text{d}F^*_t}{\text{d}t}) ||_2
\end{equation}

The total matting loss $\mathcal{L}^{M}$ is:
\vspace{-5pt}
\begin{equation}
    \mathcal{L}^{M} = \mathcal{L}^{\alpha}_{l1} + \mathcal{L}^{\alpha}_{lap} + 5 \mathcal{L}^{\alpha}_{tc} + \mathcal{L}^{F}_{l1} + 5 \mathcal{L}^{F}_{tc}
    \vspace{-5pt}
\end{equation}

For semantic segmentation, our network is only trained on the human category. To learn the segmentation probability $S_t$ \wrt the ground-truth binary label $S^*_t$, we compute binary cross entropy loss:
\vspace{-5pt}
\begin{equation}
    \mathcal{L}^{S} = S^*_t (-\log(S_t)) + (1-S^*_t)(-\log(1-S_t))
    \vspace{-5pt}
\end{equation}

\section{Experimental Evaluation}

\subsection{Evaluation on Composition Datasets}

We construct our benchmark by compositing each test sample from VM, D646, and AIM datasets onto 5 video and 5 image backgrounds. Every test clip has 100 frames. Image samples are applied with motion augmentation.

% The final test set consists 50, 110, 100 clips, or 5K, 11K, 10K frames, from each dataset respectively. 

We compare our approach against state-of-the-art trimap-based method (FBA \cite{fbamatting}), background-based method (BGMv2 \cite{bgmv2} with MobileNetV2 \cite{mobilenetv2} backbone), and auxiliary-free method (MODNet \cite{modnet}). To fairly compare them for fully automatic matting, FBA uses synthetic trimaps generated by dilation and erosion of semantic segmentation method DeepLabV3 \cite{deeplabv3} with ResNet101 \cite{resnet} backbone; BGMv2 only sees the first frame's ground-truth background; MODNet applies its neighbor frame smoothing trick. We attempt to re-train MODNet on our data but get worse results, potentially due to issues during the training, so MODNet uses its official weights; BGMv2 is already trained on all three datasets; FBA has not released the training code at the time of writing.

We evaluate $\alpha$ \wrt ground-truth $\alpha^*$ using MAD (mean absolute difference), MSE (mean squared error), Grad (spatial gradient) \cite{imagemattingmetrics}, and Conn (connectivity) \cite{imagemattingmetrics} for quality, and adopt dtSSD \cite{dtSSD} for temporal coherence. For $F$, we only measure pixels where $\alpha^*>0$ by MSE. MAD and MSE are scaled by $1e^3$ and dtSSD is scaled by $1e^2$ for better readability. $F$ is not measured on VM since it contains noisy ground-truth. MODNet does not predict $F$, so we evaluate on the input frame as its foreground prediction. This simulates directly applying the alpha matte on the input.

\begin{table}[h!]
    \centering
    \setlength\tabcolsep{1 pt}
    \begin{tabularx}{.48\textwidth}{ll|rrrrr|r}
        \toprule
        \multicolumn{2}{c|}{} & \multicolumn{5}{c|}{Alpha} & \multicolumn{1}{c}{FG} \\
        Dataset & Method & MAD & MSE & Grad & Conn & dtSSD & MSE \\
        
        \midrule
        
        \multirowcell{5.5}{VM \\ \tiny{512$\times$288}}
        & DeepLabV3 & 14.47 & 9.67 & 8.55 & 1.69 & 5.18 & \\
        & FBA & 8.36 & 3.37 & 2.09 & 0.75 & 2.09 &  \\
        & BGMv2 & 25.19 & 19.63 & 2.28 & 3.26 & 2.74 &  \\
        \cmidrule{2-8}
        & MODNet & 9.41 & 4.30 & 1.89 & 0.81 & 2.23 & \\
        & Ours & \textbf{6.08} & \textbf{1.47} & \textbf{0.88} & \textbf{0.41} & \textbf{1.36} & \\
        
        \midrule
        
        \multirowcell{5.5}{D646 \\ \tiny{512$\times$512}}
        & DeepLabV3 & 24.50 & 20.1 & 20.30 & 6.41 & 4.51 & \\
        & FBA & 17.98 & 13.40 & 7.74 & 4.65 & 2.36 & 5.84 \\
        & BGMv2 & 43.62 & 38.84 & 5.41 & 11.32 & 3.08 & \textbf{2.60} \\
        \cmidrule{2-8}
        & MODNet & 10.62 & 5.71 & 3.35 & 2.45 & 1.57 & 6.31 \\
        & Ours & \textbf{7.28} & \textbf{3.01} & \textbf{2.81} & \textbf{1.83} & \textbf{1.01} & 2.93 \\
        
        \midrule
        
        \multirowcell{5.5}{AIM \\ \tiny{512$\times$512}}
        & DeepLabV3 & 29.64 & 23.78 & 20.17 & 7.71 & 4.32 & \\
        & FBA & 23.45 & 17.66 & 9.05 & 6.05 & 2.29 & 6.32 \\
        & BGMv2 & 44.61 & 39.08 & 5.54 & 11.60 & 2.69 & \textbf{3.31} \\
        \cmidrule{2-8}
        & MODNet & 21.66 & 14.27 & 5.37 & 5.23 & 1.76 & 9.51 \\
        & Ours & \textbf{14.84} & \textbf{8.93} & \textbf{4.35} & \textbf{3.83} & \textbf{1.01} & 5.01\\
        
        \bottomrule
    \end{tabularx}
    \vspace{-8pt}
    \caption{Low-resolution comparison. Our alpha prediction is better than all others. Our foreground prediction is behind BGMv2 but outperforms FBA and MODNet. Note that FBA uses synthetic trimap from DeepLabV3; BGMv2 only sees ground-truth background from the first frame; MODNet does not predict foreground so it is evaluated on the input image.}
    \label{tab:eval_quantitative_lr}
    \vspace{-10pt}
\end{table}

\begin{table}[h!]
    \centering
    \setlength\tabcolsep{2 pt}
    \begin{tabularx}{.48\textwidth}{lX|rrrr}
        \toprule
        % \multicolumn{2}{c|}{} & \multicolumn{4}{c|}{Alpha} \\
        Dataset & Method & SAD & MSE & Grad & dtSSD \\
        \midrule
        \multirowcell{2}{VM\vspace{-5pt}\\\tiny{1920$\times$1080}}
        & MODNet + FGF & 11.13 & 5,54 & 15.30 & 3.08  \\
        & Ours & \textbf{6.57} & \textbf{1.93} & \textbf{10.55} & \textbf{1.90} \\
        \midrule
        \multirowcell{2}{D646\vspace{-5pt}\\\tiny{2048$\times$2048}}
        & MODNet + FGF & 11.27 & 6.13 & 30.78 & 2.19 \\
        & Ours & \textbf{8.67} & \textbf{4.28} & \textbf{30.06} & \textbf{1.64}\\
        \midrule
        \multirowcell{2}{AIM\vspace{-5pt}\\\tiny{2048$\times$2048}}
        & MODNet + FGF & 17.29 & 10.10 & 35.52 & 2.60 \\
        & Ours & \textbf{14.89} & \textbf{9.01} & \textbf{34.97} & \textbf{1.71}\\
        \bottomrule
    \end{tabularx}
    \vspace{-8pt}
    \caption{High-resolution alpha comparison. Ours is better than MODNet with Fast Guided Filter (FGF).}
    \label{tab:eval_quantitative_hr}
    \vspace{-15pt}
\end{table}

Table \ref{tab:eval_quantitative_lr} compares methods using low-resolution input. Our method does not use DGF in this scenario. Ours predicts more accurate and consistent alpha across all datasets. In particular, FBA is limited by the inaccurate synthetic trimap. BGMv2 performs poorly for dynamic backgrounds. MODNet produces less accurate and coherent results than ours. For foreground prediction, ours is behind BGMv2 but outperforms FBA and MODNet.

Table \ref{tab:eval_quantitative_hr} further compares our method with MODNet on high-resolution. Since DGF must be trained end-to-end with the network, we modify MODNet to use non-learned Fast Guided Filter (FGF) to upsample the prediction. Both methods use downsample scale $s=0.25$ for the encoder-decoder network. We remove Conn metric because it is too expansive to compute at high-resolution. Our method outperforms MODNet on all metrics.

\subsection{Evaluation on Real Videos}

\begin{figure*}[h!]
    \centering
    \setlength\tabcolsep{3pt}
    \newcolumntype{Y}{>{\centering\arraybackslash}X}

    \begin{Row}
        \begin{Cell}{1}
            \begin{subfigure}[t]{0.98\textwidth}
                \begin{scriptsize}
                \begin{tabularx}{\textwidth}{Y|Y|Y|Y|Y}
                    Input & FBA & BGMv2 & MODNet & Ours
                \end{tabularx}
                \end{scriptsize}
                \includegraphics[width=\textwidth]{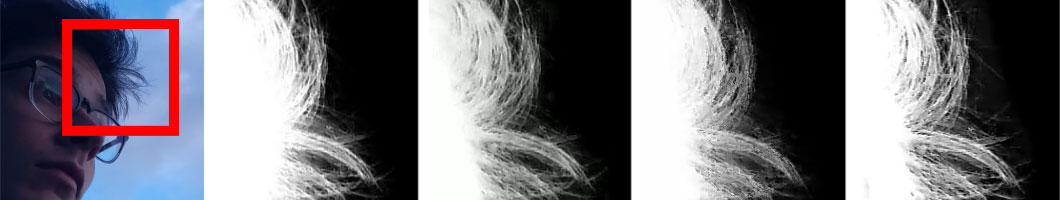}
                \caption{Alpha Detail}
                \label{fig:evaluation_qualitative_alpha}
                \vspace*{10pt}
            \end{subfigure}
            \begin{subfigure}{0.98\textwidth}
                \begin{scriptsize}
                \begin{tabularx}{\textwidth}{Y|Y|Y|Y}
                    Input & FBA & MODNet & Ours
                \end{tabularx}
                \end{scriptsize}
                \includegraphics[width=\textwidth]{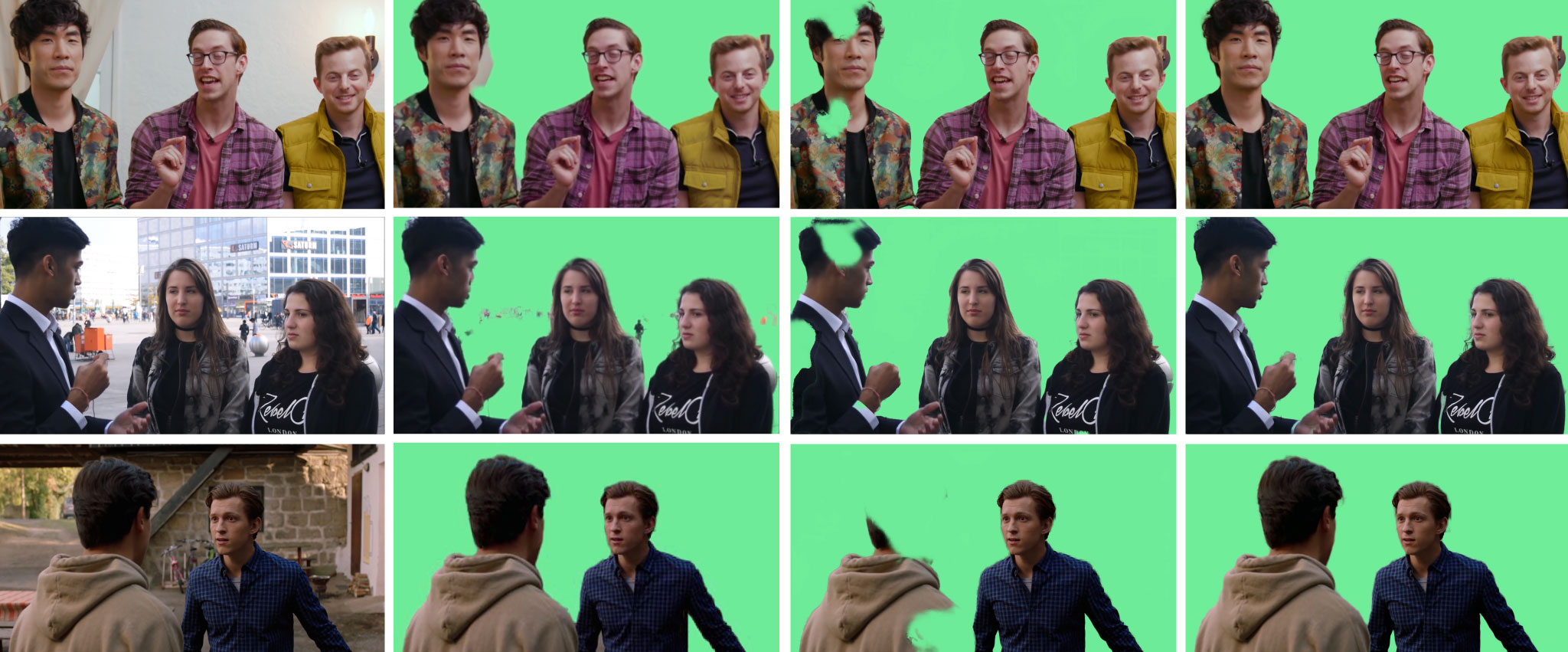}
                \caption{YouTube Videos}
                \label{fig:evaluation_qualitative_youtube}
            \end{subfigure}
        \end{Cell}
        \begin{Cell}{1}
            \begin{subfigure}[t]{0.98\textwidth}
                \begin{scriptsize}
                \begin{tabularx}{\textwidth}{Y|Y|Y|Y|Y|Y}
                    Input & MODNet & Ours & Input & MODNet & Ours
                \end{tabularx}
                \end{scriptsize}
                \includegraphics[width=\textwidth]{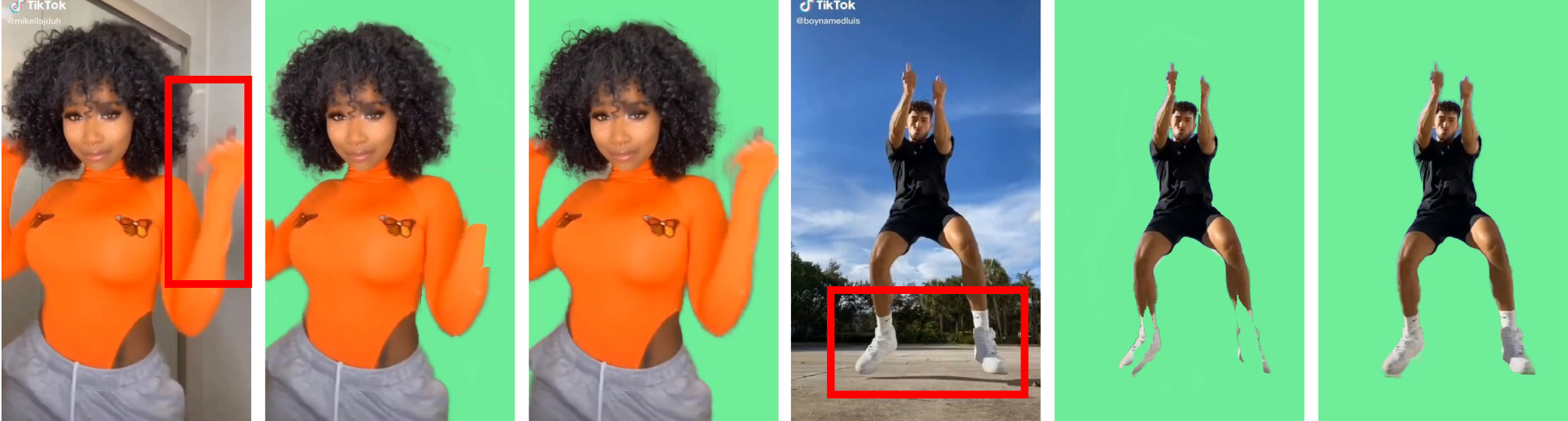}
                \caption{Cellphone Videos}
                \label{fig:evaluation_qualitative_cellphone}
                \vspace*{2pt}
            \end{subfigure}
            \begin{subfigure}{0.98\textwidth}
                \begin{scriptsize}
                \begin{tabularx}{\textwidth}{Y|Y|Y}
                    Input & MODNet & Ours
                \end{tabularx}
                \end{scriptsize}
                \includegraphics[width=\textwidth]{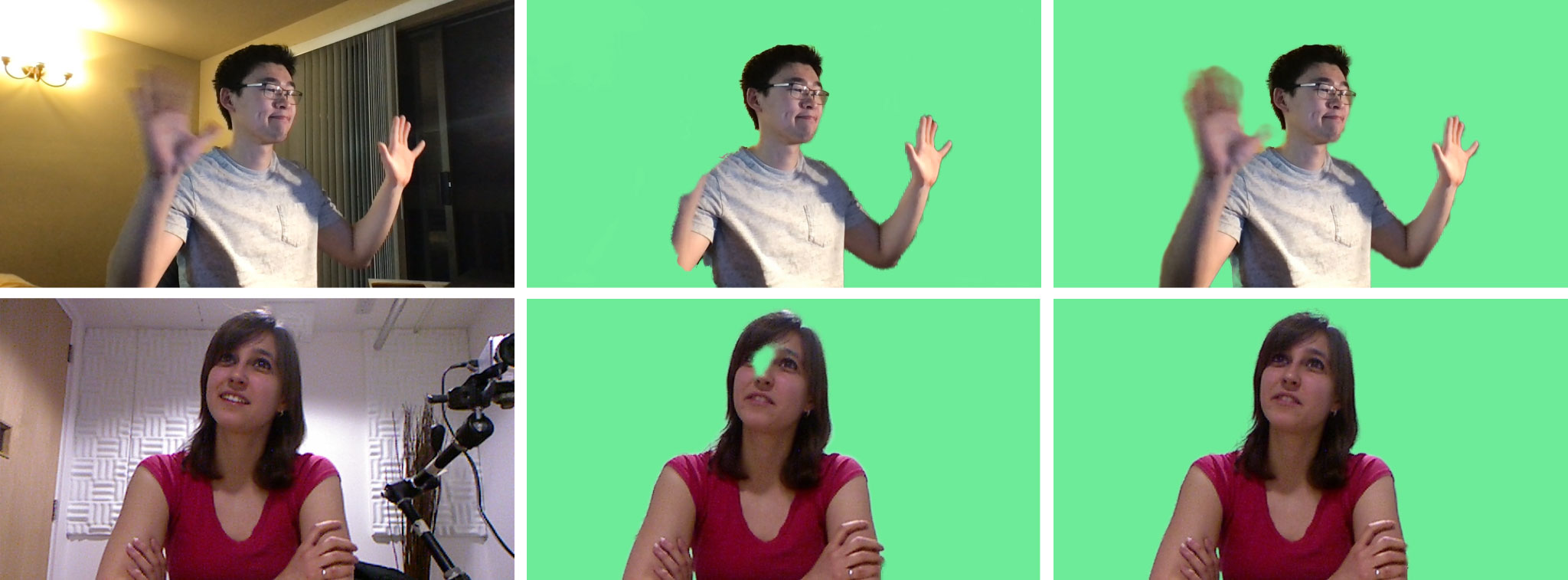}
                \caption{Webcam Videos}
                \label{fig:evaluation_qualitative_webcam}
            \end{subfigure}
        \end{Cell}
    \end{Row}
    \vspace{-8pt}
    \caption{Qualitative comparison. Our method produces more detailed alpha compared to others. When evaluating on YouTube, cellphone, and webcam videos, ours is consistently more robust than others. See supplementary for more results. YouTube videos are crawled from the Internet; cellphone videos are from a public dataset \cite{tiktokdataset}; Some webcam examples are recorded while others are taken from \cite{cam3d}.}
    \label{fig:evaluation_qualitative}
    \vspace{-10pt}
\end{figure*}

Figure \ref{fig:evaluation_qualitative} shows qualitative comparisons on real-videos. In Figure \ref{fig:evaluation_qualitative_alpha}, we compare alpha predictions across all methods and find ours predicts fine-grained details like hair strands more accurately. In Figure \ref{fig:evaluation_qualitative_youtube}, we experiment on random YouTube videos. We remove BGMv2 from the comparison since these videos do not have pre-captured backgrounds. We find our method is much more robust to semantic errors. In Figures \ref{fig:evaluation_qualitative_cellphone} and \ref{fig:evaluation_qualitative_webcam}, we further compare real-time matting against MODNet on cellphone and webcam videos. Our method can handle fast-moving body parts better than MODNet.

\subsection{Size and Speed Comparison}
Tables \ref{tab:eval_model_size} and \ref{tab:eval_model_speed} show that our method is significantly lighter, with only 58\% parameters compared to MODNet. Ours is the fastest on HD ($1920\times1080$), but a little slower than BGMv2 on $512\times288$ and MODNet with FGF on 4K ($3840\times2160$). Our inspection finds that DGF and FGF incur very minor differences in performance. Our method is slower than MODNet in 4K because ours predicts foreground in addition to alpha, so it is slower to process 3 extra channels in high resolution. We use \cite{pytorch_flopcounter} to measure GMACs (multiply–accumulate operations), but it only measures convolutions and misses out resize and many tensor operations which are used most in DGF and FGF, so GMACs is only a rough approximation. Our method achieves HD 104 FPS and 4K 76 FPS, which is considered real-time for many applications.

\begin{table}[h!]
    \centering
    \begin{tabularx}{.48\textwidth}{Xrr}
      \toprule
      Method & Parameters (Million) & Size (MB) \\
      \midrule
      DeepLabV3 & 60.996 & 233.3 \\
      FBA & 34.693 & 138.8 \\
      BGMv2 & 5.007 & 19.4 \\
      MODNet & 6.487 & 25.0 \\
      \midrule
      Ours & \textbf{3.749} & \textbf{14.5} \\
      \bottomrule
    \end{tabularx}
    \vspace{-8pt}
    \caption{Ours is lighter than all compared methods. Size is measured on FP32 weights.}
    \label{tab:eval_model_size}
    \vspace{-10pt}
\end{table}

\begin{table}[h!]
    \centering
    \setlength\tabcolsep{2 pt}
    \begin{tabularx}{.48\textwidth}{ccXrr}
        \toprule
        Resolution & $s$ & Method & FPS & GMACs$^*$ \\
        \midrule
        \multirow{4}{*}{512$\times$288} & 
        \multirow{4}{*}{1}
          & DeepLabV3 + FBA & 12.3 & 205.77 \\
        & & BGMv2 & \textbf{152.5} & 8.46 \\
        & & MODNet & 104.9 & 8.80 \\
        & & Ours & 131.9 & \textbf{4.57} \\
        \midrule
        \multirow{3}{*}{1920$\times$1080} &
        \multirow{3}{*}{0.25}
          & BGMv2 & 70.6 & 9.86 \\
        & & MODNet + FGF & 100.3 & 7.78 \\
        & & Ours & \textbf{104.2} & \textbf{4.15} \\
        \midrule
        \multirow{3}{*}{3840$\times$2160} &
        \multirow{3}{*}{0.125}
          & BGMv2 & 26.5 & 17.04 \\
        & & MODNet + FGF & \textbf{88.6} & 7.78 \\
        & & Ours & 76.5 & \textbf{4.15} \\
        \bottomrule
    \end{tabularx}
    \vspace{-8pt}
    \caption{Model performance comparison. $s$ denotes the downsample scale. Models are converted to TorchScript and optimized before testing (BatchNorm fusion \etc). FPS is measured as FP32 tensor throughput on an Nvidia GTX 1080Ti GPU. GMACs is a rough approximation.}
    \label{tab:eval_model_speed}
    \vspace{-10pt}
\end{table}

\section{Ablation Studies}

\subsection{Role of Temporal Information}

\begin{figure}[h!]
    \centering
    \includegraphics[width=0.48\textwidth]{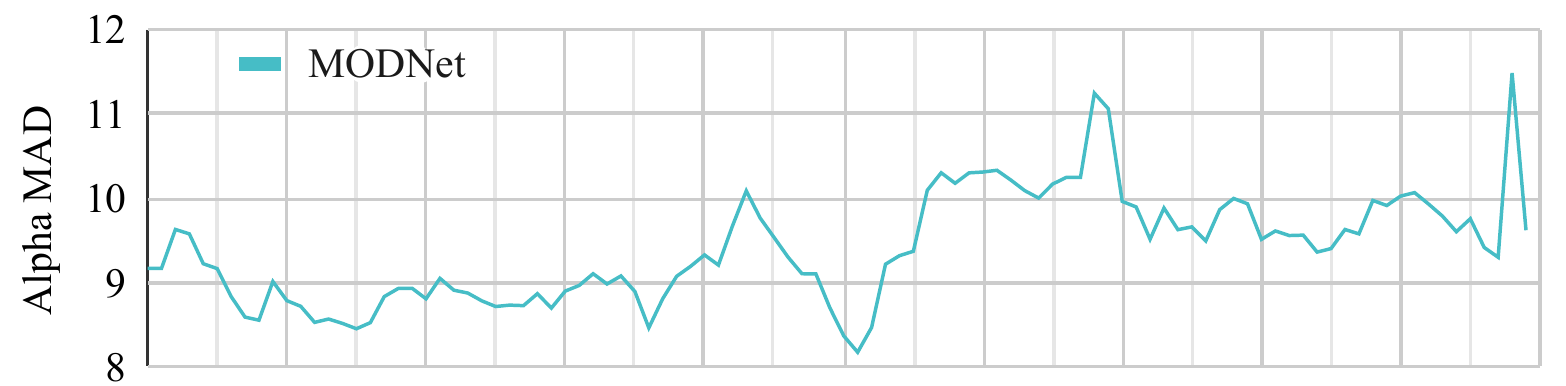}
    \includegraphics[width=0.48\textwidth]{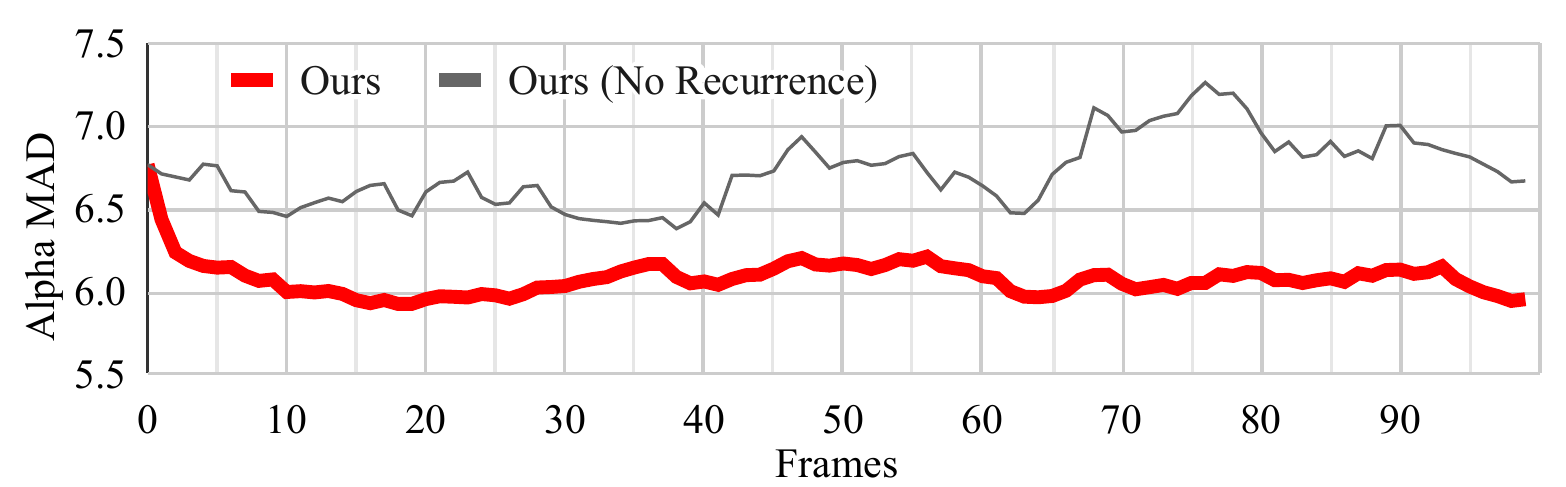}
    \vspace{-20pt}
    \caption{Average alpha MAD over time on VM without DGF. Our metric improves over time and is stable, showing that temporal information improves quality and consistency.}
    \label{fig:ablation_temporal_chart}
    \vspace{-10pt}
\end{figure}

Figure \ref{fig:ablation_temporal_chart} shows the change of average alpha MAD metric across all VM test clips over time. The error of our model drops significantly in the first 15 frames then the metric stays stable. MODNet, even with its neighbor frame smoothing trick, has large fluctuations in the metric. We also experiment with disabling recurrence in our network by passing zero tensors as the recurrent states. The quality and consistency worsen as expected. This proves that temporal information improves quality and consistency.

\begin{figure}[h!]
    \centering
    \newcolumntype{Y}{>{\centering\arraybackslash}X}
    \setlength\tabcolsep{0pt}
    \begin{scriptsize}
    \begin{tabularx}{.45\textwidth}{cY|Y|Y|Y}
         & $t$ & $t + 1$ & $t + 2$ & $t + 3$ \\
        \rotatebox{90}{\parbox{38pt}{\centering{Input}}} & \multicolumn{4}{c}{\multirow[b]{3}{*}[28pt]{\includegraphics[width=.43\textwidth]{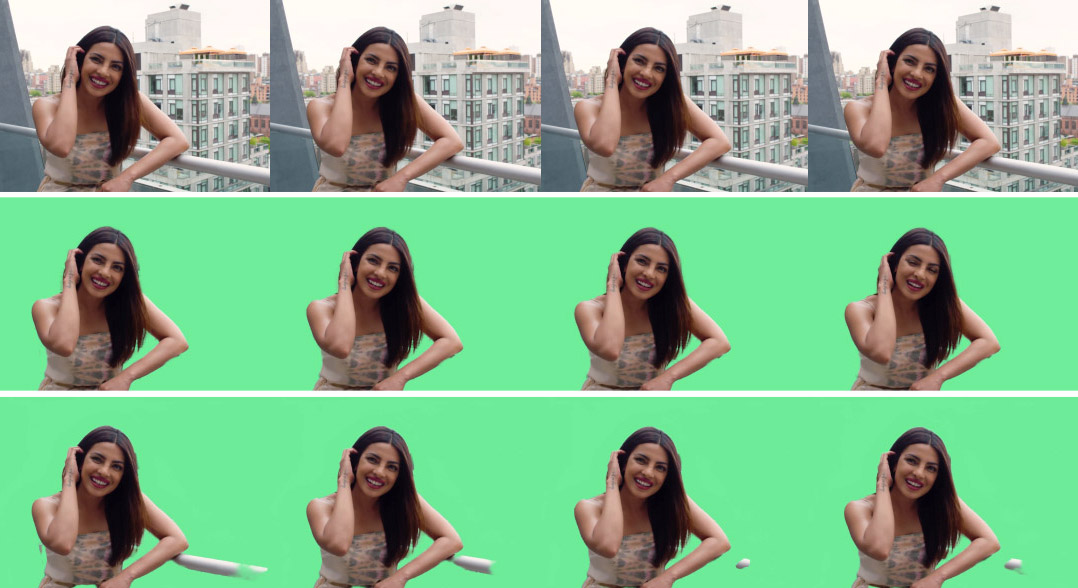}}} \\
        \cline{0-0}
        \rotatebox{90}{\parbox{38pt}{\centering{Ours}}}\\
        \cline{0-0}
        \rotatebox{90}{\parbox{38pt}{\centering{MODNet}}}\\
    \end{tabularx}
    \end{scriptsize}
    \vspace{-8pt}
    \caption{Temporal coherence comparison. MODNet's result has flicker on the handrail while ours is consistent.}
    \label{fig:ablation_temporal_coherence}
    \vspace{-10pt}
\end{figure}

Figure \ref{fig:ablation_temporal_coherence} compares temporal coherence with MODNet on a video sample. Our method produces consistent results on the handrail region while MODNet produces flicker, which significantly degrade perceptual quality. Please see our supplementary for more results.

\begin{figure}[h!]
    \centering
    \newcolumntype{Y}{>{\centering\arraybackslash}X}
    \setlength\tabcolsep{0pt}
    \begin{scriptsize}
    \begin{tabularx}{.45\textwidth}{cY|Y|Y|Y}
         & $t=0$ & $t=100$ & $t=200$ & $t=300$ \\
        \rotatebox{90}{\parbox{38pt}{\centering{Input}}} & \multicolumn{4}{c}{\multirow[b]{2}{*}[28pt]{\includegraphics[width=.43\textwidth]{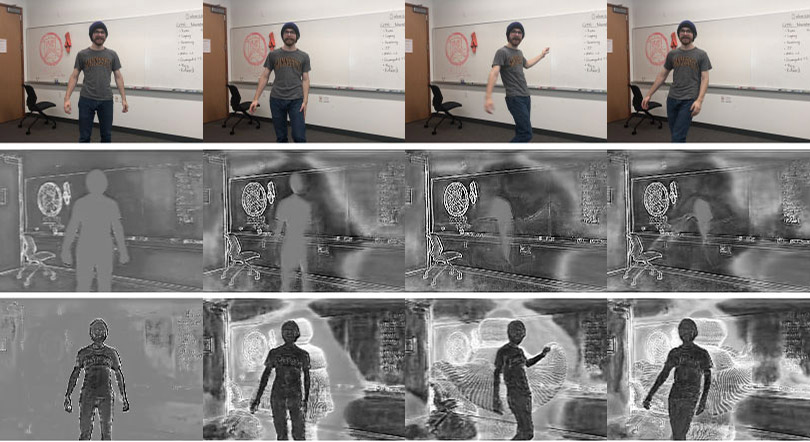}}} \\
        \cline{0-0}
        \rotatebox{90}{\parbox{76pt}{\centering{Recurrent Channels}}}\\
    \end{tabularx}
    \end{scriptsize}
    
    \vspace{-5pt}
    \caption{Two example channels in the recurrent hidden state. Our network learns to to reconstruct the background over time and keeps track of the motion history in its recurrent state.}
    \label{fig:ablation_temporal_hidden}
    \vspace{-10pt}
\end{figure}

We further examine the recurrent hidden state. In Figure \ref{fig:ablation_temporal_hidden}, we find that our network has automatically learned to reconstruct the background as it is revealed over time and keep this information in its recurrent channels to help future predictions. It also uses other recurrent channels to keep track of motion history. Our method even attempts to reconstruct the background when the videos contain camera movements and is capable of forgetting useless memory on shot cuts. More examples are in the supplementary.

\subsection{Role of Segmentation Training Objective}
Table \ref{tab:ablation_segmentation_metrics} shows that our method is as robust as the semantic segmentation methods when evaluated on the subset of COCO validation images that contain humans and only on the human category. Our method achieves 61.50 mIOU, which is reasonably in between the performance of MobileNetV3 and DeepLabV3 trained on COCO considering the difference in model size. We also try to evaluate the robustness of our alpha output by thresholding $\alpha > 0.5$ as the binary mask and our method still achieves 60.88 mIOU, showing that the alpha prediction is also robust. For comparison, we train a separate model by initializing our MobileNetV3 encoder and LR-ASPP module with the pre-trained weights on COCO and removing the segmentation objective. The model overfits to the synthetic matting data and regresses significantly on COCO performance, achieving only 38.24 mIOU.

\vspace{-3pt}
\begin{table}[!h]
    \centering
    \begin{tabularx}{.48\textwidth}{Xr}
      \toprule
      Method &  mIOU \\
      \midrule
      DeepLabV3 & 68.93 \\
      MobileNetV3 + LR-ASPP & 58.58 \\
      \midrule
      Ours (alpha output, no seg objective) & 38.24 \\
      \midrule
      Ours (alpha output) & \textbf{60.88} \\
      Ours (segmentation output) & \textbf{61.50} \\
      \bottomrule
    \end{tabularx}
    \vspace{-5pt}
    \caption{Segmentation performance on COCO validation set. Training with segmentation objective makes our method robust while training only with pre-trained weights regresses.}
    \label{tab:ablation_segmentation_metrics}
    \vspace{-10pt}
\end{table}

\subsection{Role of Deep Guided Filter}\label{sec:ablation_dgf}
Table \ref{tab:ablation_dgf} shows that DGF has only a small overhead in size and speed compared to FGF. DGF has a better Grad metric, indicating its high-resolution details are more accurate. DGF also produces more coherent results indicated by the dtSSD metric, likely because it takes hidden features from the recurrent decoder into consideration. The MAD and MSE metrics are inconclusive because they are dominated by segmentation-level errors, which are not corrected by either DGF or FGF.

\begin{table}[!h]
    \centering
    \setlength\tabcolsep{3 pt}
    \begin{tabularx}{.48\textwidth}{X|rr|rrrr}
      \toprule
      Method & Params & FPS & MAD & MSE & Grad & dtSSD \\
      \midrule
      Ours (FGF) & \textbf{3.748} & \textbf{109.4} & 8.70 & \textbf{4.13} & 31.44 & 1.89\\
      Ours & 3.749 & 104.2 & \textbf{8.67} & 4.28 & \textbf{30.06} & \textbf{1.64} \\
      \bottomrule
    \end{tabularx}
    \vspace{-5pt}
    \caption{Comparing switching DGF to FGF on D646. Parameters are measured in millions. FPS is measured in HD.}
    \label{tab:ablation_dgf}
    \vspace{-10pt}
\end{table}

\subsection{Static vs. Dynamic Backgrounds}

Table \ref{tab:ablation_background_type} compares the performance on static and dynamic backgrounds. Dynamic backgrounds include both background object movements and camera movements. Our method can handle both cases and performs slightly better on static backgrounds, likely because it is easier to reconstruct pixel-aligned backgrounds as shown in Figure \ref{fig:ablation_temporal_hidden}. On the other hand, BGMv2 performs badly on dynamic backgrounds and MODNet does not exhibit any preference. In metric, BGMv2 outperforms ours on static backgrounds, but it is expected to do worse in reality when the pre-captured background has misalignment.

\begin{table}[!h]
    \centering
    \setlength\tabcolsep{2 pt}
    \begin{tabularx}{.48\textwidth}{lX|rrrr}
      \toprule
      Background & Method & MAD & MSE & Grad & dtSSD \\
      \midrule
      \multirow{3}{*}{Static}
       & BGMv2$^*$ & \textbf{4.33} & \textbf{0.32} & \textbf{4.19} & \textbf{1.33} \\
       & MODNet + FGF & 11.04 & 5.42 & 15.80 & 3.10\\ 
       & Ours & 5.64 & 1.07 & 9.80 & 1.84\\
      \midrule
      \multirow{3}{*}{Dynamic}
       & BGMv2 & 42.45 & 37.05 & 17.30 & 4.61\\
       & MODNet + FGF & 11.23 & 5.65 & 14.79 & 3.06 \\ 
       & Ours & \textbf{7.50} & \textbf{2.80} & \textbf{11.30} & \textbf{1.96} \\
      \bottomrule
    \end{tabularx}
    \vspace{-5pt}
    \caption{Comparing VM samples on static and dynamic backgrounds. Ours does better on static backgrounds but can handle both cases. Note that BGMv2 receives ground-truth static backgrounds, but in reality the backgrounds have misalignment.}
    \label{tab:ablation_background_type}
    \vspace{-10pt}
\end{table}

\subsection{Larger Model for Extra Performance}
We experiment with switching the backbone to ResNet50 \cite{resnet} and increasing the decoder channels. Table \ref{tab:ablation_larger} shows the performance improvement. The large model is more suitable for server-side applications.

% We experiment with switching the backbone to ResNet50 \cite{resnet} and increasing the decoder channels. Table \ref{tab:ablation_larger} shows that the large model achieves better performance and can still reach HD 71 FPS. The large model has significantly more parameters so it is more suitable for server-side applications.

\begin{table}[!h]
    \centering
    \setlength\tabcolsep{1.5 pt}
    \begin{tabularx}{.48\textwidth}{X|rrr|rrrr}
      \toprule
      Method & Params & Size & FPS & MAD & MSE & Grad & dtSSD \\
      \midrule
      Ours Large & 26.890 & 102.9 & 71.1 & \textbf{5.81} & \textbf{0.97} & \textbf{9.65} & \textbf{1.78} \\
      Ours & \textbf{3.749} & \textbf{14.5} & \textbf{104.2} & 6.57 & 1.93 & 10.55 & 1.90 \\
      \bottomrule
    \end{tabularx}
    \vspace{-5pt}
    \caption{Large model uses ResNet50 backbone and has more decoder channels. Evaluated on VM in HD. Size is measured in MB.}
    \label{tab:ablation_larger}
    \vspace{-15pt}
\end{table}

\subsection{Limitations}

\begin{figure}[h!]
    \centering
    \includegraphics[width=0.48\textwidth]{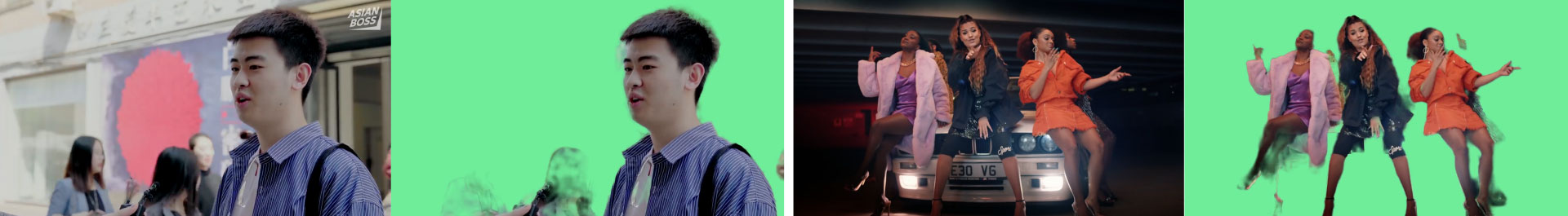}
    \vspace{-15pt}
    \caption{Challenging cases. People in the background make matting target ambiguous. Complex scenes make matting harder.}
    \label{fig:ablation_limitations}
    \vspace{-10pt}
\end{figure}

Our method prefers videos with clear target subjects. When there are people in the background, the subjects of interest become ambiguous. It also favors simpler backgrounds to produce more accurate matting. Figure \ref{fig:ablation_limitations} shows examples of challenging cases.

\section{Conclusion}

We have proposed a recurrent architecture for robust human video matting. Our method achieves new state-of-the-art while being lighter and faster. Our analysis shows that temporal information plays an important role in improving the quality and consistency. We also introduce a new training strategy to train our model on both matting and semantic segmentation objectives. This approach effectively enforces our model to be robust on various types of videos.

\clearpage
\newpage

{\small
\bibliographystyle{ieee_fullname}
\bibliography{bib}
}

\clearpage
\newpage

\renewcommand{\thesection}{\Alph{section}}
\setcounter{section}{0}

\section{Overview} 

We provide additional details in this supplementary. In Section \ref{sec:supp_model}, we describe the details of our network architecture. In Section \ref{sec:supp_training}, we explain the details on training. In Section \ref{sec:supp_data}, we show examples of our composited matting data samples. In Section \ref{sec:supp_results}, we show additional results from our method. We also attach video results in the supplementary. Please see our videos for better visualization.
\section{Network}
\label{sec:supp_model}

\begin{table}[h!]
    \centering
    \setlength\tabcolsep{1.5pt}
    \begin{tabularx}{.48\textwidth}{l|rrrr|r|rrrrr}
        \toprule
        Backbone & $E_\frac{1}{2}$ & $E_\frac{1}{4}$ & $E_\frac{1}{8}$ & $E_\frac{1}{16}$ & $AS$ & $D_\frac{1}{16}$ & $D_\frac{1}{8}$ & $D_\frac{1}{4}$ & $D_\frac{1}{2}$ & $D_\frac{1}{1}$ \\
        \midrule
        % MB-Small & 16 & 16 & 24 & 576 & 128 & 128 & 64 & 32 & 24 & 16\\
        \textbf{Ours} & \textbf{16} & \textbf{24} & \textbf{40} & \textbf{960} & \textbf{128} & \textbf{128} & \textbf{80} & \textbf{40} & \textbf{32} & \textbf{16}\\
        Ours Large & 64 & 256 & 512 & 2048 & 256 & 256 & 128 & 64 & 32 & 16\\
        \bottomrule
    \end{tabularx}
    \vspace{-5pt}
    \caption{Feature channels at different scale. $E_k$ and $D_k$ denote encoder and decoder channels at $k$ feature scale respectively. $AS$ denotes LR-ASPP channels.}
    \label{tab:supp_model_channels}
    \vspace{-10pt}
\end{table}

Table \ref{tab:supp_model_channels} describes our network and its variants with feature channels. Our default network uses MobileNetV3-Large \cite{mobilenetv3} backbone while the large variant uses ResNet50 \cite{resnet} backbone.

\textbf{Encoder:} The encoder backbone operates on individual frames and extracts feature maps of $E_k$ channels at $k \in [\frac{1}{2}, \frac{1}{4}, \frac{1}{8}, \frac{1}{16}]$ scales. Unlike regular MobileNetV3 and ResNet backbones that continue to operate at $\frac{1}{32}$ scale, we modify the last block to use convolutions with a dilation rate of 2 and a stride of 1 following the design of \cite{deeplabv3,deeplabv3+,mobilenetv3}. The last feature map $E_\frac{1}{16}$ is given to the LR-ASPP module, which compresses it to $AS$ channels.

\textbf{Decoder:} All ConvGRU layers operate on half of the channels by split and concatenation, so the recurrent hidden state has $\frac{D_k}{2}$ channels at scale $k$. For the upsampling blocks, the convolution, Batch Normalization, and ReLU stack compresses the concatenated features to $D_k$ channels before splitting to ConvGRU. For the output block, the first two convolutions have 16 filters and the final hidden features has 16 channels. The final projection convolution outputs 5 channels, including 3-channel foreground, 1-channel alpha, and 1-channel segmentation predictions. All convolutions uses $3\times3$ kernels except the last projection uses a $1\times1$ kernel. The average poolings use $2\times2$ kernels with a stride of 2.

\textbf{Deep Guided Filter:} DGF contains a few $1\times1$ convolutions internally. We modify it to take the predicted foreground, alpha, and the final hidden features as inputs. All internal convolutions use 16 filters. Please refer to \cite{deepguidedfilter} for more specifications. 

Our entire network is built and trained in PyTorch \cite{pytorch}. We clamp the alpha and foreground prediction outputs to $[0, 1]$ range without activation functions following \cite{fbamatting,bgmv2}. The clamp is done during both training and inference. The segmentation prediction output is sigmoid logits.
\section{Training}
\label{sec:supp_training}

\begin{algorithm}[h!]
    \For{stage $\in [1, 2, 3, 4]$}{
        \For{epoch}{
            \For{iteration}{
                LowResMattingPass($B, T, h, w$)
                
                \If{stage $\in [3, 4]$}{
                    HighResMattingPass($B, \hat{T}, \hat{h}, \hat{w}$)
                }
                \eIf{iteration \% 2 = 0}{
                    VideoSegmentationPass($B, T, h, w$)
                }{
                    ImageSegmentationPass($B', 1, h, w$)
                }
            }
        }
    }
    \caption{Training Procedures}
    \label{algo:supp_training_algorithm}
\end{algorithm}

Algorithm \ref{algo:supp_training_algorithm} shows the training loop of our proposed training strategy. The sequence length parameters $T$, $\hat{T}$ are set according to the stages, which is specified in our main text; the batch size parameters are set to $B=4$, and $B'=B \times T$; The input resolutions are randomly sampled as $h, w \sim Uniform(256, 512)$ and $\hat{h}, \hat{w} \sim Uniform(1024, 2048)$.

Our network is trained using 4 Nvidia V100 32G GPUs. We use mixed precision training \cite{mixedprecisiontraining} to reduce the GPU memory consumption. The training takes approximately 18, 2, 8, and 14 hours in each stage respectively.
\section{Data Samples}
\label{sec:supp_data}

\begin{figure}[h!]
    \centering
    \setlength\tabcolsep{0pt}
    \newcolumntype{Y}{>{\centering\arraybackslash}X}
    \newcolumntype{P}[1]{>{\centering\arraybackslash}p{#1}}
    \begin{scriptsize}
        \begin{tabularx}{0.48\textwidth}{|Y|P{.096\textwidth}|}
            Composited Frames & Std Dev
        \end{tabularx}
    \end{scriptsize}
    \includegraphics[width=.48\textwidth]{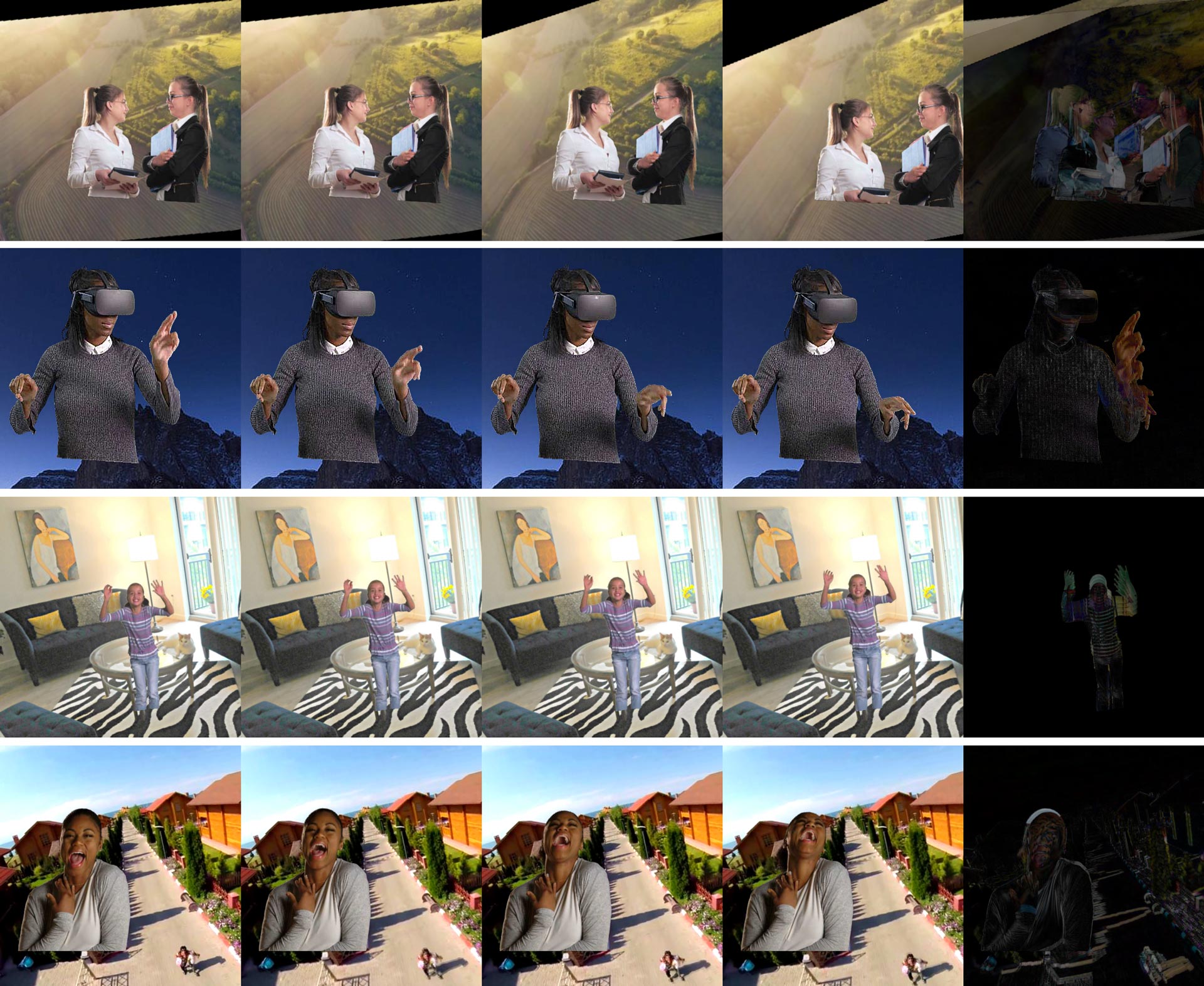}
    \vspace{-15pt}
    \caption{Composited training samples. Last column shows the standard deviation of each pixel across time to visualize motion.}
    \label{fig:supp_data_train_samples}
    \vspace{-5pt}
\end{figure}

Figure \ref{fig:supp_data_train_samples} shows examples of composited training samples from the matting datasets. The clips contain natural movements when compositing with videos as well as artificial movements generated by the motion augmentation.

\begin{figure}[h!]
    \centering
    \includegraphics[width=.48\textwidth]{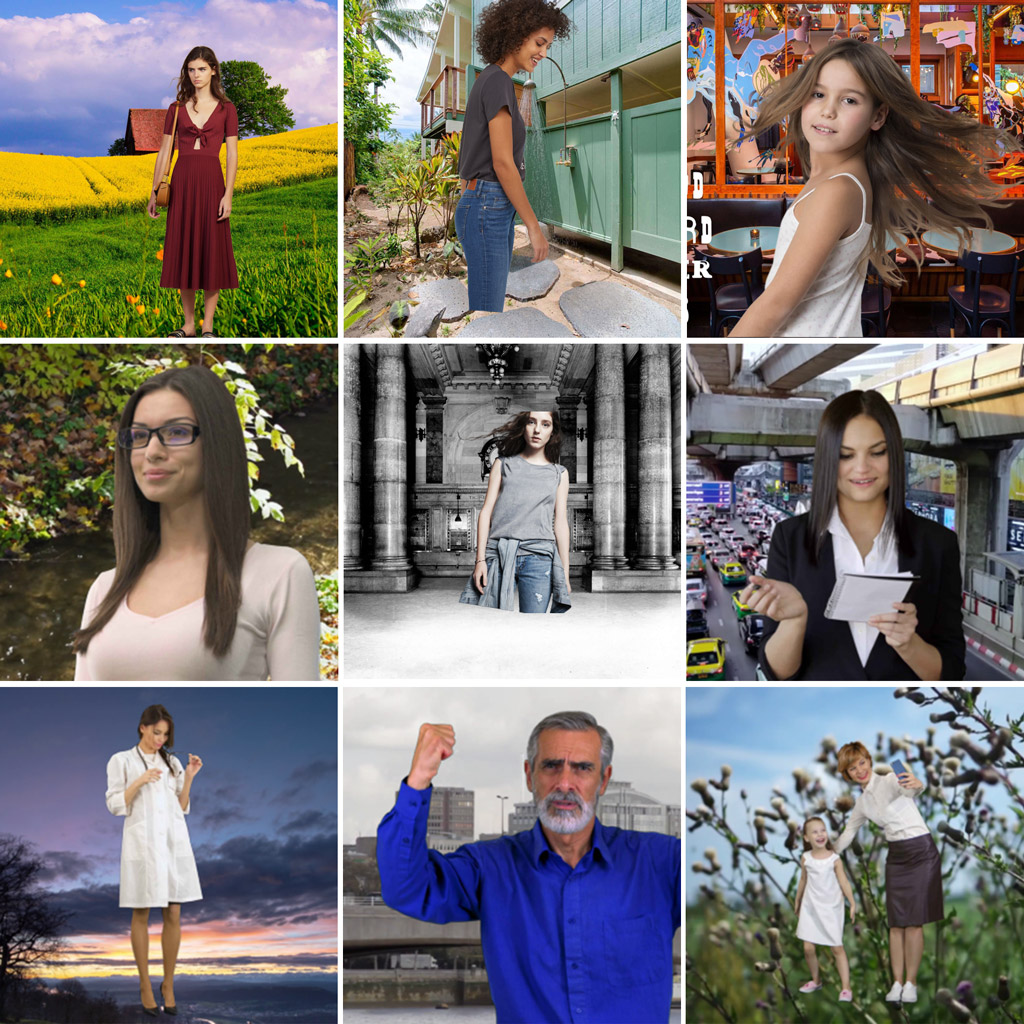}
    \vspace{-15pt}
    \caption{Example testing samples. The augmentation is only applied on image foreground and background. The augmentation strength is weaker to make samples look more realistic.}
    \label{fig:supp_data_test_samples}
    \vspace{-8pt}
\end{figure}

Figure \ref{fig:supp_data_test_samples} shows examples of the composited testing samples. The testing samples only apply motion augmentation on image foreground and backgrounds. The motion augmentation only consists of affine transforms. The strength of the augmentation is also weaker compared to the training augmentation to make testing samples as realistic looking as possible.

\section{Additional Results}
\label{sec:supp_results}

\begin{figure*}
    \centering
    \newcolumntype{Y}{>{\centering\arraybackslash}X}
    \begin{scriptsize}
    \begin{tabularx}{.98\textwidth}{Y|Y|Y|Y|Y|Y}
        Input & Ours & MODNet & Input & Ours & MODNet
    \end{tabularx}
    \end{scriptsize}
    \includegraphics[width=.98\textwidth]{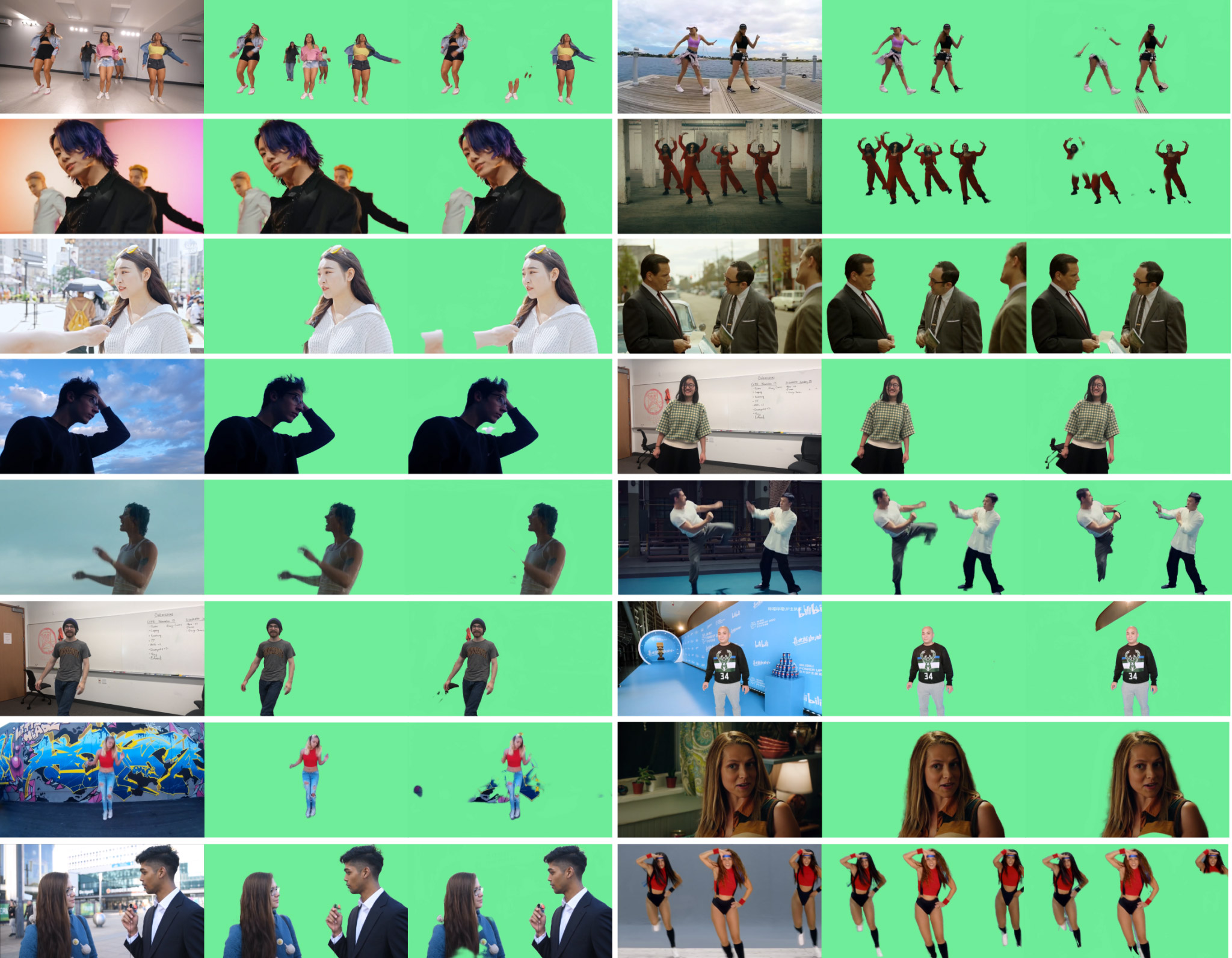}
    \caption{More qualitative comparison with MODNet.}
    \label{fig:supp_results_quality}
\end{figure*}

\begin{figure*}
    \centering
    \newcolumntype{Y}{>{\centering\arraybackslash}X}
    \setlength\tabcolsep{0pt}
    \begin{scriptsize}
    \begin{tabularx}{\textwidth}{cY|Y|Y|Y|Y|Y}
         & $t$ & $t + 1$ & $t + 2$ & $t + 3$ & $t + 4$ & $t + 5$ \\
        \rotatebox{90}{\parbox{45pt}{\centering{Input}}} & \multicolumn{6}{c}{\multirow[b]{3}{*}[36pt]{\includegraphics[width=.98\textwidth]{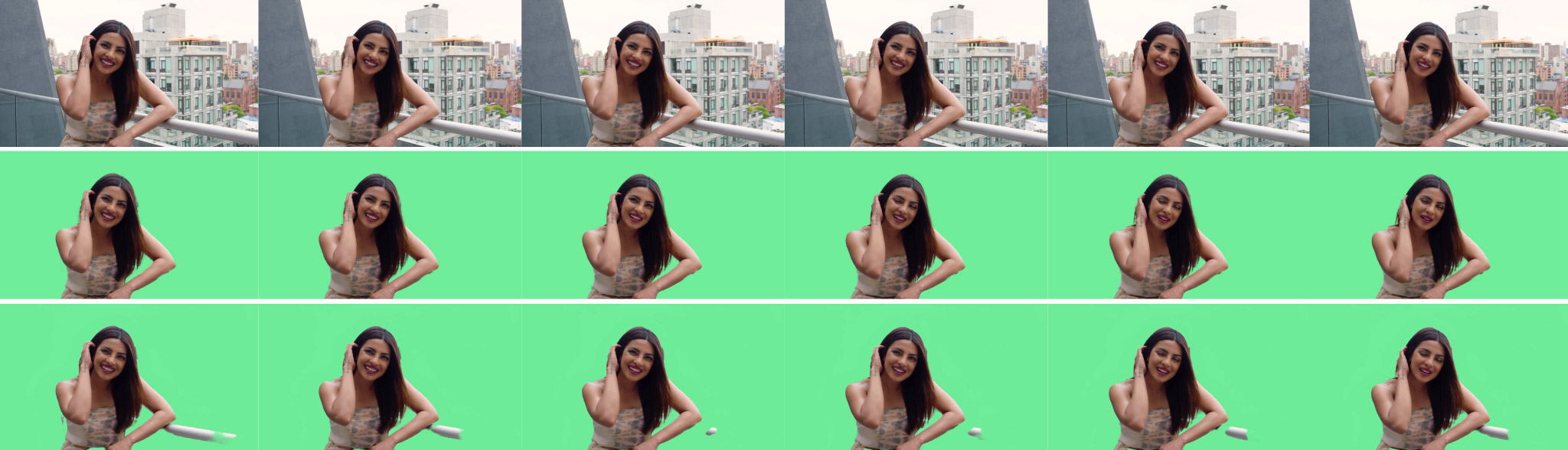}}} \\
        \cline{0-0}
        \rotatebox{90}{\parbox{45pt}{\centering{Ours}}}\\
        \cline{0-0}
        \rotatebox{90}{\parbox{45pt}{\centering{MODNet}}}\\
    \end{tabularx}
    \end{scriptsize}
    \vspace{-8pt}
    \caption{Temporal coherence comparison. Our result is temporally coherent, whereas MODNet produces flicker around the handrail. This is because MODNet processes every frame as indepdendent images, so its matting decision is not consistent.}
    \label{fig:supp_results_coherence}
\end{figure*}

\begin{figure*}
    \centering

    \newcolumntype{Y}{>{\centering\arraybackslash}X}
    \setlength\tabcolsep{0pt}
    \begin{subfigure}{0.98\textwidth}
        \begin{scriptsize}
        \begin{tabularx}{\textwidth}{cY|Y|Y|Y|Y|Y}
             & $t=0$ & $t=100$ & $t=200$ & $t=300$ & $t=400$ & $t=500$ \\
            \rotatebox{90}{\parbox{45pt}{\centering{Input}}} & \multicolumn{6}{c}{\multirow[b]{2}{*}[36pt]{\includegraphics[width=.98\textwidth]{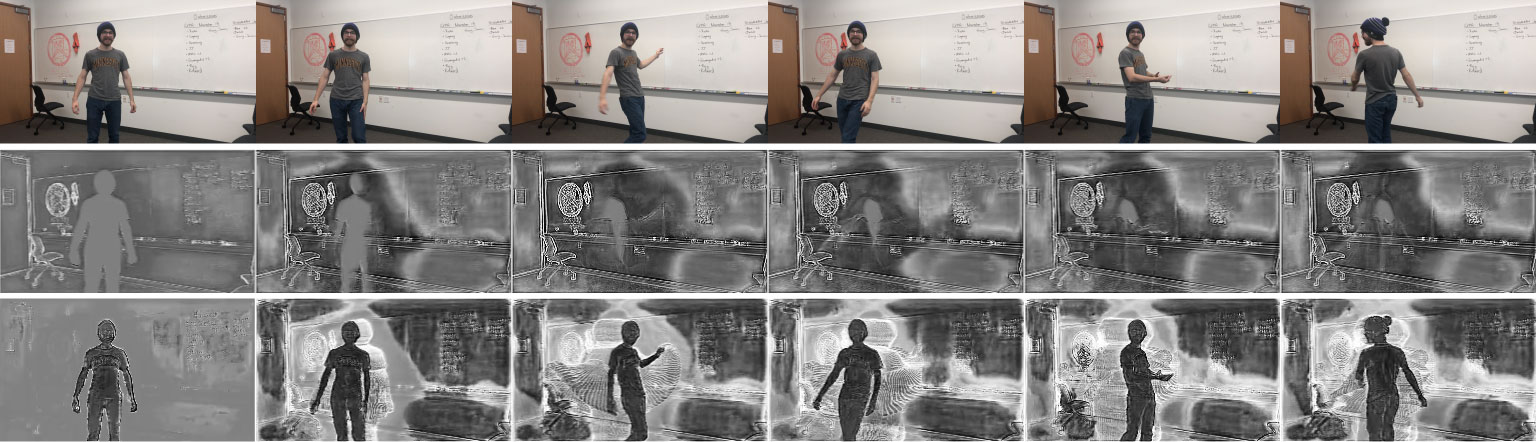}}} \\
            \cline{0-0}
            \rotatebox{90}{\parbox{90pt}{\centering{Recurrent Channels}}}\\
        \end{tabularx}
        \end{scriptsize}
        \vspace{-5pt}
        \caption{Video with a static background.}
        \vspace{5pt}
    \end{subfigure}
    \begin{subfigure}{0.98\textwidth}
        \begin{scriptsize}
        \begin{tabularx}{\textwidth}{cY|Y|Y|Y|Y|Y}
             & $t=0$ & $t=100$ & $t=200$ & $t=300$ & $t=400$ & $t=500$ \\
            \rotatebox{90}{\parbox{45pt}{\centering{Input}}} & \multicolumn{6}{c}{\multirow[b]{2}{*}[36pt]{\includegraphics[width=.98\textwidth]{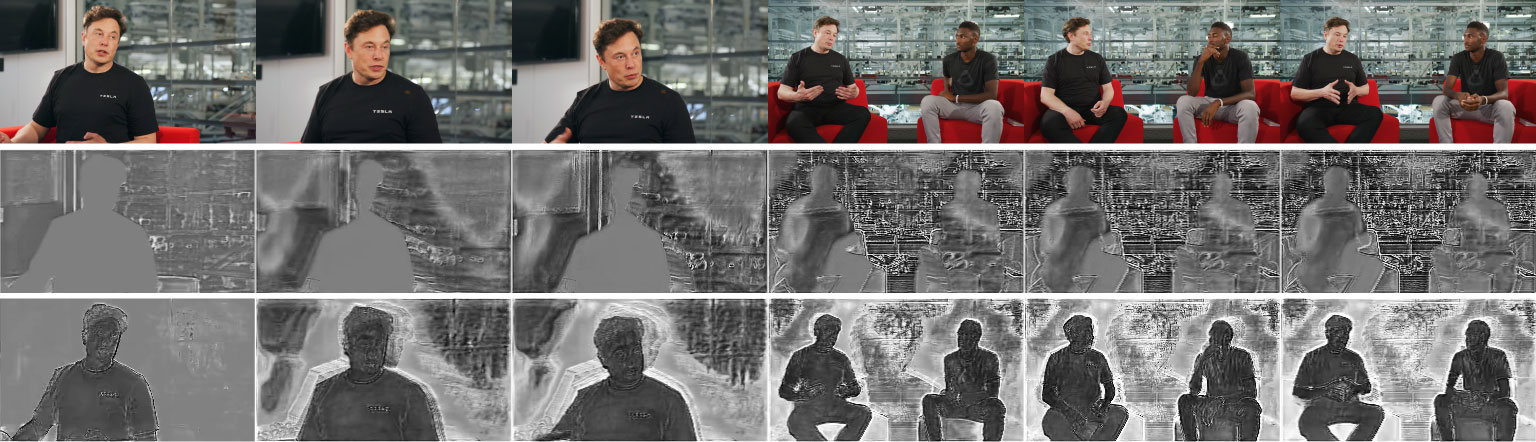}}} \\
            \cline{0-0}
            \rotatebox{90}{\parbox{90pt}{\centering{Recurrent Channels}}}\\
        \end{tabularx}
        \end{scriptsize}
        \vspace{-5pt}
        \caption{Video with a handheld camera and cut shots.}
        \vspace{-5pt}
    \end{subfigure}
    \caption{More examples of the recurrent hidden states. The first example with the static background clearly shows our model reconstructs the occluded background region over time. The second example with a handheld camera shows that our model still attempts to reconstruct the background, and it has learned to forget useless recurrent states on shot cuts.}
    \label{fig:supp_results_hidden}
\end{figure*}

Figure \ref{fig:supp_results_quality} shows additional qualitative comparisons with MODNet. Our method is consistently more robust. Figure \ref{fig:supp_results_coherence} compares temporal coherence with MODNet. MODNet has flicker on low-confidence regions whereas our results are coherent. Figure \ref{fig:supp_results_hidden} shows additional examples of our model's recurrent hidden state. It shows that our model has learned to store useful temporal information in its recurrent state and is capable of forgetting useless information upon shot cuts.

\end{document}